\documentclass[10pt,twocolumn,letterpaper]{article}

\usepackage{cvpr}
\usepackage{times}
\usepackage{epsfig}
\usepackage{graphicx}
\usepackage{amsmath}
\usepackage{amssymb}
\usepackage[boxed]{algorithm}
\usepackage{algorithmic}

\newcommand{\MDBN}[1]{\text{CGBN}(#1)}

\usepackage{subfigure}
\usepackage{caption}
\usepackage{comment}
\usepackage{multirow}

\usepackage{paralist}

\newcommand{\mmAR}{$\text{mmAR}$}

\usepackage[pagebackref=true,breaklinks=true,letterpaper=true,colorlinks,bookmarks=false]{hyperref}

\cvprfinalcopy 

\def\cvprPaperID{836} 

\author{
    Chao Peng\thanks{Equal contribution. This work is done when Zeming Li and Tete Xiao are interns at Megvii Research.} \quad Tete Xiao$^{1*}$ \quad Zeming Li$^{2*}$ \quad Yuning Jiang \quad Xiangyu Zhang \quad Kai Jia \quad Gang Yu \quad Jian Sun \vspace{0.10cm} \\
    $^1$Peking University, jasonhsiao97@pku.edu.cn\\
    $^2$Tsinghua University, lizm15@mails.tsinghua.edu.cn\\
    Megvii Inc. (Face++), \{pengchao, jyn, zhangxiangyu, jiakai, yugang, sunjian\}@megvii.com
}

\begin{document}

\title{MegDet: A Large Mini-Batch Object Detector}

\maketitle

\begin{abstract}   
The development of object detection in the era of deep learning, from R-CNN~\cite{girshick2014rich}, Fast/Faster R-CNN~\cite{girshick2015fast,ren2015faster} to recent Mask R-CNN~\cite{He_2017_ICCV} and RetinaNet~\cite{Lin_2017_ICCV}, mainly come from novel network, new framework, or loss design. However, mini-batch size, a key factor for the training of deep neural networks, has not been well studied for object detection. In this paper, we propose a Large Mini-Batch Object Detector (MegDet) to enable the training with a large mini-batch size up to 256, so that we can effectively utilize at most 128 GPUs to significantly shorten the training time. Technically, we suggest a warmup learning rate policy and Cross-GPU Batch Normalization, which together allow us to successfully train a large mini-batch detector in much less time (e.g., from 33 hours to 4 hours), and achieve even better accuracy. The MegDet is the backbone of our submission (mmAP \textbf{52.5\%}) to COCO 2017 Challenge, where we won the \textbf{1st} place of Detection task.
\end{abstract}

\section{Introduction}

\begin{figure}[t]
      \begin{center}
         \includegraphics[width=1\linewidth]{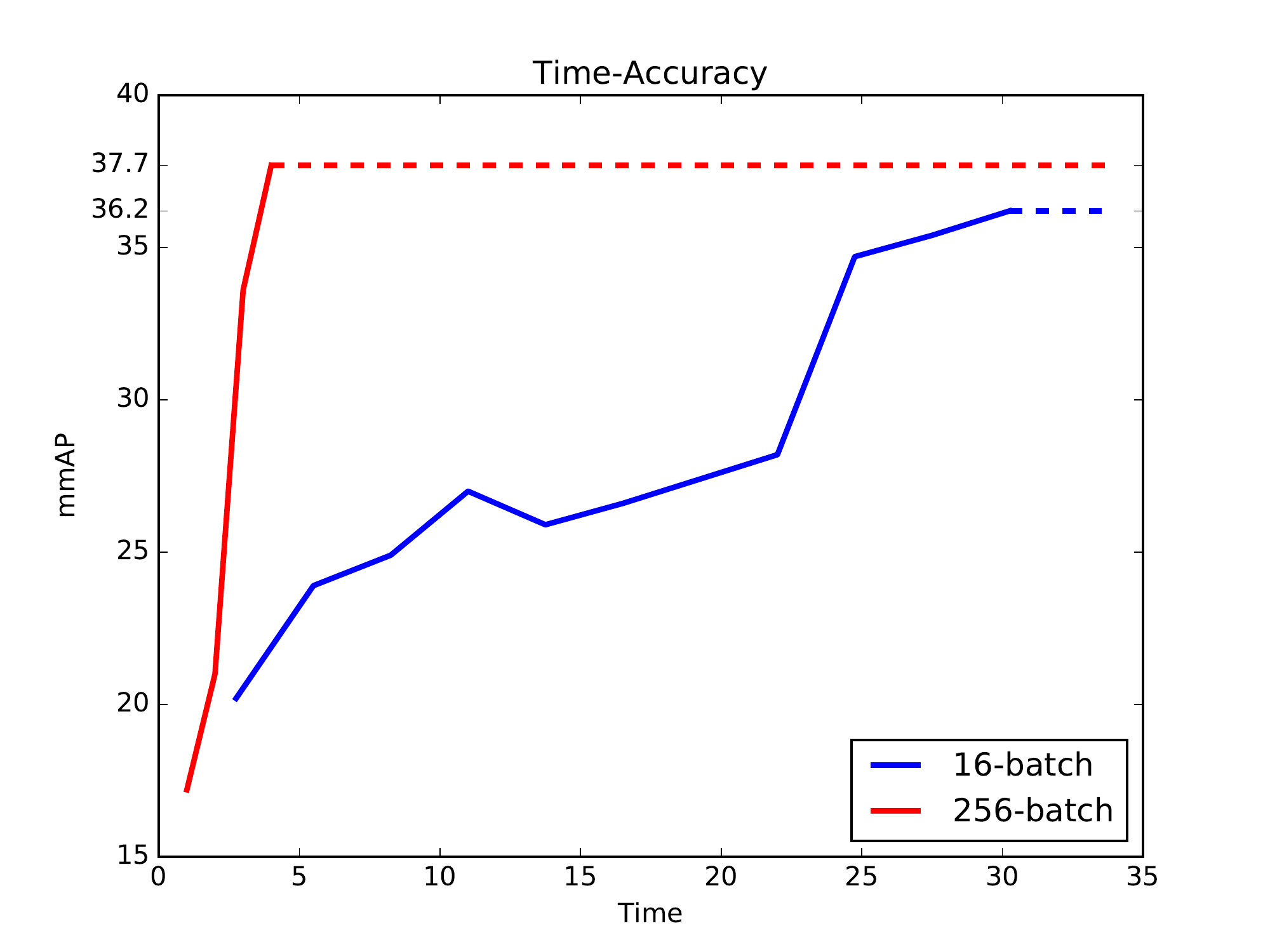}
      \end{center}
      \caption{Validation accuracy of the same FPN object detector trained on COCO dataset, with mini-batch size 16 (on 8 GPUs) and mini-batch size 256 (on 128 GPUs). The large mini-batch detector is more accurate and its training is nearly an order-of-magnitude faster.}
      \label{fig:figure_1}
    \end{figure}
    
Tremendous progresses have been made on CNN-based object detection, since seminal work of R-CNN~\cite{girshick2014rich}, Fast/Faster R-CNN series~\cite{girshick2015fast,ren2015faster}, and recent state-of-the-art detectors like Mask R-CNN~\cite{He_2017_ICCV} and RetinaNet~\cite{Lin_2017_ICCV}. Taking COCO~\cite{lin2014microsoft} dataset as an example, its performance has been boosted from $19.7$ AP in Fast R-CNN~\cite{girshick2015fast} to $39.1$ AP in RetinaNet~\cite{Lin_2017_ICCV}, in just two years. The improvements are mainly due to better backbone network~\cite{he2016deep}, new detection framework~\cite{ren2015faster}, novel loss design~\cite{Lin_2017_ICCV},  improved pooling method~\cite{dai2016instance,He_2017_ICCV}, and so on~\cite{huang2016speed}. 

A recent trend on CNN-based image classification uses very large min-batch size to significantly speed up the training. For example, the training of ResNet-50 can be accomplished in an hour~\cite{goyal2017accurate}  or even in 31 minutes~\cite{you2017100} , using mini-batch size 8,192 or 16,000, with little or small sacrifice on the accuracy. In contract, the mini-batch size remains very small (e.g., 2-16) in object detection literatures. Therefore in this paper, we study the problem of mini-batch size in object detection and present a technical solution to successfully train a large mini-batch size object detector. 

\emph{What is wrong with the small mini-batch size}? Originating from the object detector R-CNN series, a mini-batch involving only $2$ images is widely adopted in popular detectors like  Faster R-CNN and R-FCN. Though in state-of-the-art detectors like RetinaNet and Mask R-CNN the mini-batch size is increased to $16$, which is still quite small compared with the mini-batch size (e.g., 256) used in current image classification. There are several potential drawbacks associated with small mini-batch size. First, the training time is notoriously lengthy. For example, the training of ResNet-152 on COCO takes 3 days, using the mini-bath size 16 on a machine with 8 Titian XP GPUs. Second, training with small mini-batch size fails to provide accurate statistics for batch normalization~\cite{ioffe2015batch} (BN). In order to obtain a good batch normalization statistics, the mini-batch size for ImageNet classification network is usually set to 256, which is significantly larger than the mini-batch size used in current object detector setting. 

\begin{figure}
	    \centering
	    \includegraphics[width=0.8\linewidth]{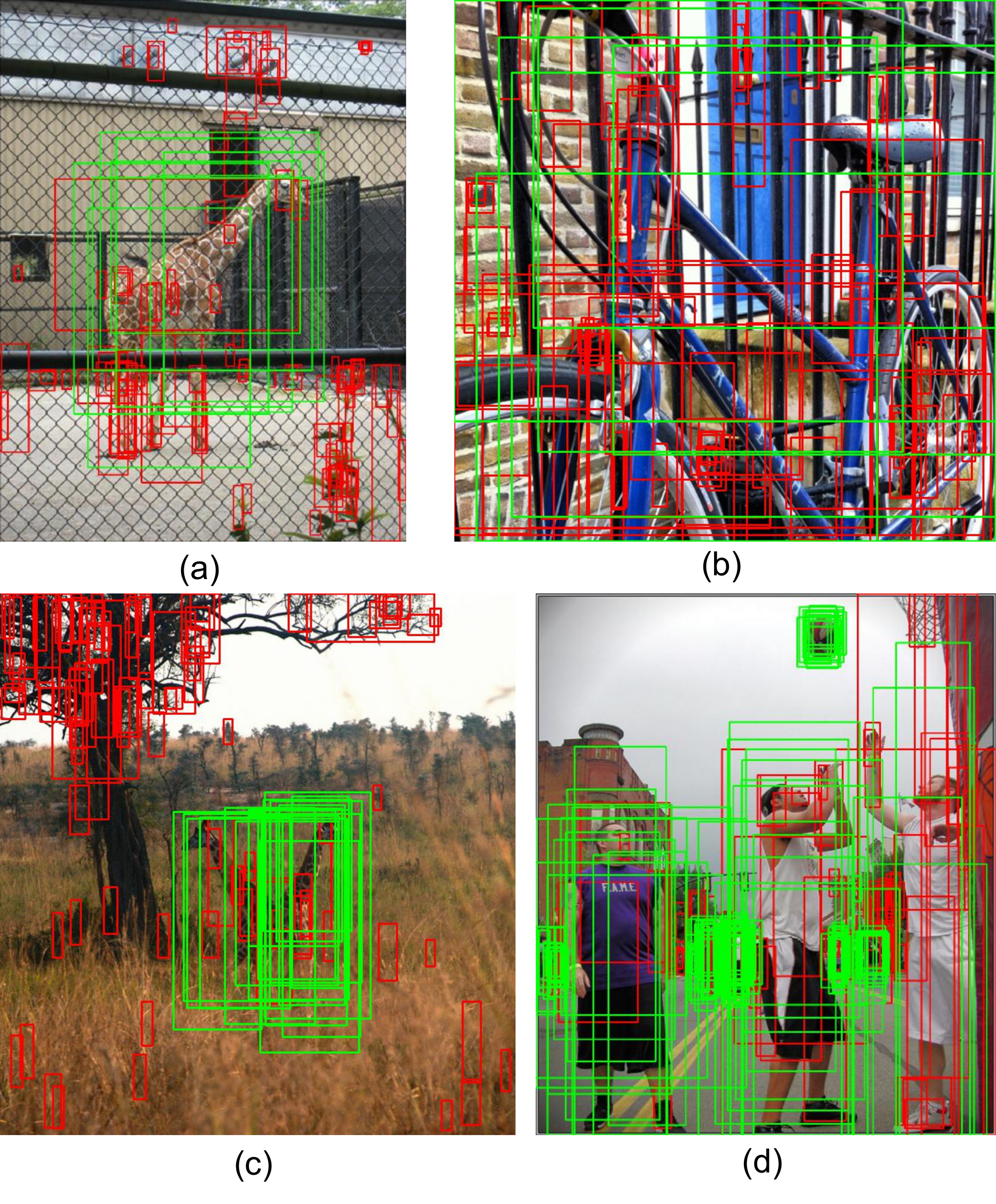}
	    \caption{Example images with positive and negative proposals. (a-b) two examples with imbalanced ratio, (c-d) two examples with moderate balanced ratio. Note that we sub-sampled the negative proposals for visualization.}
	    \label{fig:rois_example}
	\end{figure}

Last but not the least, the number of positive and negative training examples within a small mini-batch are more likely imbalanced, which might hurt the final accuracy. Figure~\ref{fig:rois_example} gives some examples with imbalanced positive and negative proposals. And Table~\ref{tab:pos_neg_ratio} compares the statistics of two detectors with different mini-batch sizes, at different training epochs on COCO dataset. 

\emph{What is the challenge to simply increase the min-batch size?} As in the image classification problem, the main dilemma we are facing is: the large min-batch size usually requires a large learning rate to maintain the accuracy, according to ``equivalent learning rate rule"~\cite{goyal2017accurate,krizhevsky2014one}. But a large learning rate in object detection could be very likely leading to the failure of convergence; if we use a smaller learning rate to ensure the convergence, an inferior results are often obtained.

To tackle the above dilemma, we propose a solution as follows. First, we present a new explanation of linear scaling rule and borrow the ``warmup" learning rate policy~\cite{goyal2017accurate} to gradually increase the learning rate at the very early stage. This ensures the convergence of training. Second, to address the accuracy and convergence issues, we introduce Cross-GPU Batch Normalization (CGBN) for better BN statistics. CGBN not only improves the accuracy but also makes the training much more stable. This is significant because we are able to safely enjoy the rapidly increased computational power from industry.

Our MegDet (ResNet-50 as backbone) can finish COCO training in ~4 hours on 128 GPUs, reaching even higher accuracy. In contrast, the small mini-batch counterpart takes 33 hours with lower accuracy. This means that we can speed up the innovation cycle by nearly an order-of-magnitude with even better performance, as shown in Figure ~\ref{fig:figure_1}. Based on MegDet, we secured \textbf{1st} place of COCO 2017 Detection Challenge.

Our technical contributions can be summarized as:

\begin{itemize}
\item We give a new interpretation of linear scaling rule, in the context of object detection, based on an assumption of maintaining equivalent loss variance.
\item We are the first to train BN in the object detection framework. We demonstrate that our Cross-GPU Batch Normalization not only benefits the accuracy, but also makes the training easy to converge, especially for the large mini-batch size. 
\item We are the first to finish the COCO training (based on ResNet-50) in 4 hours, using 128 GPUs, and achieving higher accuracy.
\item Our MegDet leads to the winning of COCO 2017 Detection Challenge.
\end{itemize}

        \begin{table}[t]
    		\begin{center}
    			\begin{tabular}{|c|c|c|c|}
    			\hline
    			Epoch & Batch Size& Ratio(\%)\\
    			\hline
    			\multirow{2}{*}{1} & 16 & 5.58 \\
    			& 256 & \textbf{9.82} \\
    			\hline
    			\multirow{2}{*}{6} & 16 & 11.77 \\
    			& 256 & \textbf{16.11} \\
    			\hline
    			\multirow{2}{*}{12} & 16 & 16.59 \\
    			& 256 & \textbf{16.91} \\
    			\hline
    			\end{tabular}
    		\end{center}
    		\caption{Ratio of positive and negative samples in the training (at epoch 1, 6, 12). The larger mini-batch size makes the ratio more balanced, especially at the early stage.}
    		\label{tab:pos_neg_ratio}
    	\end{table}

\section{Related Work}
\textbf{CNN-based detectors} have been the mainstream in current academia and industry. We can roughly divide existing CNN-based detectors into two categories: one-stage detectors like SSD~\cite{liu2016ssd}, YOLO~\cite{redmon2016you,redmon2016yolo9000} and recent Retina-Net~\cite{Lin_2017_ICCV}, and two-stage detectors~\cite{sermanet2013overfeat,bell2016inside} like Faster R-CNN~\cite{ren2015faster}, R-FCN~\cite{dai2016r} and Mask-RCNN~\cite{He_2017_ICCV}. 

For two-stage detectors, let us start from the R-CNN family. R-CNN~\cite{girshick2014rich} was first introduced in 2014. It employs Selective Search~\cite{uijlings2013selective} to generate a set of region proposals and then classifies the warped patches through a CNN recognition model. As the computation of the warp process is intensive, SPPNet~\cite{he2014spatial} improves the R-CNN by performing classification on the pooled feature maps based on a spatial pyramid pooling rather than classifying on the resized raw images. Fast-RCNN~\cite{girshick2015fast} simplifies the Spatial Pyramid Pooling (SPP) to ROIPooling. Although reasonable performance has been obtained based on Fast-RCNN, it still replies on traditional methods like selective search to generate proposals. Faster-RCNN~\cite{ren2015faster} replaces the traditional region proposal method with the Region Proposal Network (RPN), and proposes an end-to-end detection framework. The computational cost of Faster-RCNN will increase dramatically if the number of proposals is large. In R-FCN~\cite{dai2016r}, position-sensitive pooling is introduced to obtain a speed-accuracy trade-off. Recent works are more focusing on improving detection performance. Deformable ConvNets~\cite{dai2017deformable} uses the learned offsets to convolve different locations of feature maps, and forces the networks to focus on the objects. FPN~\cite{Lin_2017_CVPR} introduces the feature pyramid technique and makes significant progress on small object detection. As FPN provides a good trade-off between accuracy and implementation, we use it as the default detection framework. To address the alignment issue, Mask R-CNN~\cite{He_2017_ICCV} introduces the ROIAlign and achieves state-of-the-art results for both object detection and instance segmentation.

Different from two-stage detectors, which involve a proposal and refining step, one-stage detectors usually run faster. In YOLO~\cite{redmon2016you,redmon2016yolo9000}, a convolutional network is followed with a fully connected layer to obtain classification and regression results based on a $7\times7$ grid. SSD~\cite{liu2016ssd} presents a fully convolutional network with different feature layers targeting different anchor scales. Recently, RetinaNet is introduced in~\cite{Lin_2017_ICCV} based on the focal loss, which can significantly reduce false positives in one-stage detectors.

\textbf{Large mini-batch training} has been an active research topic in image classification. In~\cite{goyal2017accurate}, imagenet training based on ResNet50 can be finished in one hour. ~\cite{you2017100} presents a training setting which can finish the ResNet50 training in 31 minutes without losing classification accuracy. Besides the training speed, ~\cite{hoffer2017train} investigates the generalization gap between large mini-batch and small mini-batch, and propose the novel model and algorithm to eliminate the gap. However, the topic of large mini-batch training for object detection is rarely discussed so far.

\section{Approach}

In this section, we present our Large Mini-Batch Detector (MegDet), to finish the training in less time while achieving higher accuracy. 

\subsection{Problems with Small Mini-Batch Size}
The early generation of CNN-based detectors use very small mini-batch size like $2$ in Faster-RCNN and R-FCN. Even in state-of-the-art detectors like RetinaNet and Mask R-CNN, the batch size is set as $16$. There exist a few problems when training with a small mini-batch size. First, we have to pay much longer training time if a small mini-batch size is utilized for training. As shown in Figure~\ref{fig:figure_1}, the training of a ResNet-50 detector based on a mini-batch size of 16 takes more than 30 hours. With the original mini-batch size $2$, the training time could be more than one week. Second, in the training of detector, we usually fix the statistics of Batch Normalization and use the pre-computed values on ImageNet dataset, since the small mini-batch size is not applicable to re-train the BN layers. It is a sub-optimal tradeoff since the two datasets, COCO and ImageNet, are much different. Last but not the least, the ratio of positive and negative samples could be very imbalanced. In Table~\ref{tab:pos_neg_ratio}, we provide the statistics for the ratio of positive and negative training examples. We can see that a small mini-batch size leads to more imbalanced training examples, especially at the initial stage. This imbalance may affect the overall detection performance. 

As we discussed in the introduction, simply increasing the mini-batch size has to deal with the tradeoff between convergence and accuracy. To address this issue, we first discuss the learning rate policy for the large mini-batch. 

\subsection{Learning Rate for Large Mini-Batch}
\label{sec:lr_for_lb}
The learning rate policy is strongly related to the SGD algorithm. Therefore, we start the discussion by first reviewing the structure of loss for object detection network, 
    \begin{align}
        \label{eq:loss_all}
        \begin{split}
            L(x, w)&= \frac{1}{N} \sum_{i=1}^N l(x_i, w) + \frac{\lambda}{2}||w||_2^2 \\
                &= l(x, w) + l(w),
        \end{split}
    \end{align}
where $N$ is the min-batch size, $l(x, w)$ is the task specific loss and $l(w)$ is the regularization loss. For Faster R-CNN~\cite{ren2015faster} framework and its variants~\cite{dai2016r,Lin_2017_CVPR,He_2017_ICCV}, $l(x_i,w)$ consists of RPN prediction loss, RPN bounding-box regression loss, prediction loss, and bounding box regression loss. 

According to the definition of mini-batch SGD, the training system needs to compute the gradients with respect to weights $w$, and updates them after every iteration. When the size of mini-batch changes, such as $\hat N \leftarrow k \cdot N$, we expect that the learning rate $r$ should also be adapted to maintain the efficiency of training. Previous works~\cite{krizhevsky2014one, goyal2017accurate,you2017100} use \emph{Linear Scaling Rule}, which changes the new learning rate to $\hat r \leftarrow k\cdot r$. Since one step in large mini-batch $\hat N$ should match the effectiveness of $k$ accumulative steps in small mini-batch $N$, the learning rate $r$ should be also multiplied by the same ratio $k$ to counteract the scaling factor in loss. This is based on a \emph{gradient equivalence} assumption \cite{goyal2017accurate} in the SGD updates. This rule of thumb has been well-verified in image classification, and we find it is still applicable for object detection. However, the interpretation is is different for a weaker and better assumption. 

In image classification, every image has only one annotation and $l(x, w)$ is a simple form of cross-entropy. As for object detection, every image has different number of box annotations, resulting in different ground-truth distribution among images. Considering the differences between two tasks, the assumption of gradient equivalence between different mini-batch sizes might be less likely to be hold in object detection. So, we introduce another explanation based on the following variance analysis.

\noindent \textbf{Variance Equivalence.} Different from the gradient equivalence assumption, we assume that the variance of gradient remain the same during $k$ steps. Given the mini-batch size $N$, if the gradient of each sample $\nabla l(x_i, w)$ obeying i.i.d., the variance of gradient on $l(x, w)$ is:
    \begin{align}
        \begin{split}
            \text{Var}(\nabla l(x, w_t)) &= \frac{1}{N^2}\sum_{i=1}^N \text{Var}(\frac{\partial l(x_i, w_t)}{\partial w_t}) \\
            & = \frac{1}{N^2} \times \left(N \cdot \sigma^2_l \right )  \\
            &= \frac{1}{N} \sigma^2_l.
        \end{split}
        \label{eq:var_result_task_loss_small_batch}
    \end{align}
    Similarly, for the large mini-batch $\hat N = k \cdot N$, we can get the following expression:
    \begin{align}
        \begin{split}
            \text{Var}(\nabla l_{\hat N}(x, w_t)) &= \frac{1}{kN} \sigma^2_l.
        \end{split}
        \label{eq:var_result_task_loss_large_batch}
    \end{align}
Instead of expecting equivalence on weight update, here we want to \emph{maintain the variance of one update in large mini-batch $\hat N$ equal to $k$ accumulative steps in small mini-batch $N$}. To achieve this, we have:
    \begin{align}
        \label{eq:eq_batch_size}
        \begin{split}
            \text{Var}(r \cdot \sum_{t=1}^k(\nabla l_{N}^t(x, w)) ) &= r^2\cdot k\cdot \text{Var}(\nabla l_N(x, w)) \\
            &\approx \hat r^2 \text{Var}(\nabla l_{\hat N}(x, w))
        \end{split}
    \end{align}
Within Equation~\eqref{eq:var_result_task_loss_small_batch} and~\eqref{eq:var_result_task_loss_large_batch}, the above equality holds if and only if $\hat r = k\cdot r$, which gives the same linear scaling rule for $\hat r$. 

Although the final scaling rule is the same, our \emph{variance equivalence} assumption on Equation~\eqref{eq:eq_batch_size}  is weaker because we just expect that the large mini-batch training can maintain equivalent statistics on the gradients. We hope the variance analysis here can shed light on deeper understanding of learning rate in wider applications. 


\noindent \textbf{Warmup Strategy. }As discussed in \cite{goyal2017accurate}, the linear scaling rule may not be applicable at the initial stage of the training, because the weights changing are dramatic. To address this practical issue, we borrow \emph{Linear Gradual Warmup} in~\cite{goyal2017accurate}. That is, we set up the learning rate small enough at the beginning, such as $r$. Then, we increase the learning rate with a constant speed after every iteration, until to $\hat r$.  

The warmup strategy can help the convergence. But as we demonstrated in the experiments later, it is not enough for larger mini-batch size, e.g., 128 or 256. Next, we introduce the Cross-GPU Batch Normalization, which is the main workhorse of large mini-batch training. 

\subsection{Cross-GPU Batch Normalization}
\label{sec:multi_device_bn}
Batch Normalization~\cite{ioffe2015batch} is an important technique for training a very deep convolutional neural network. Without batch normalization, training such a deep network will consume much more time or even fail to converge. However, previous object detection frameworks, such as FPN~\cite{Lin_2017_CVPR}, initialize models with an ImageNet pre-trained model, after which the batch normalization layer is fixed during the whole fine-tuning procedure. In this work, we make an attempt to perform batch normalization for object detection.

\begin{figure}[t]
  \begin{center}
     \includegraphics[width=0.6\linewidth]{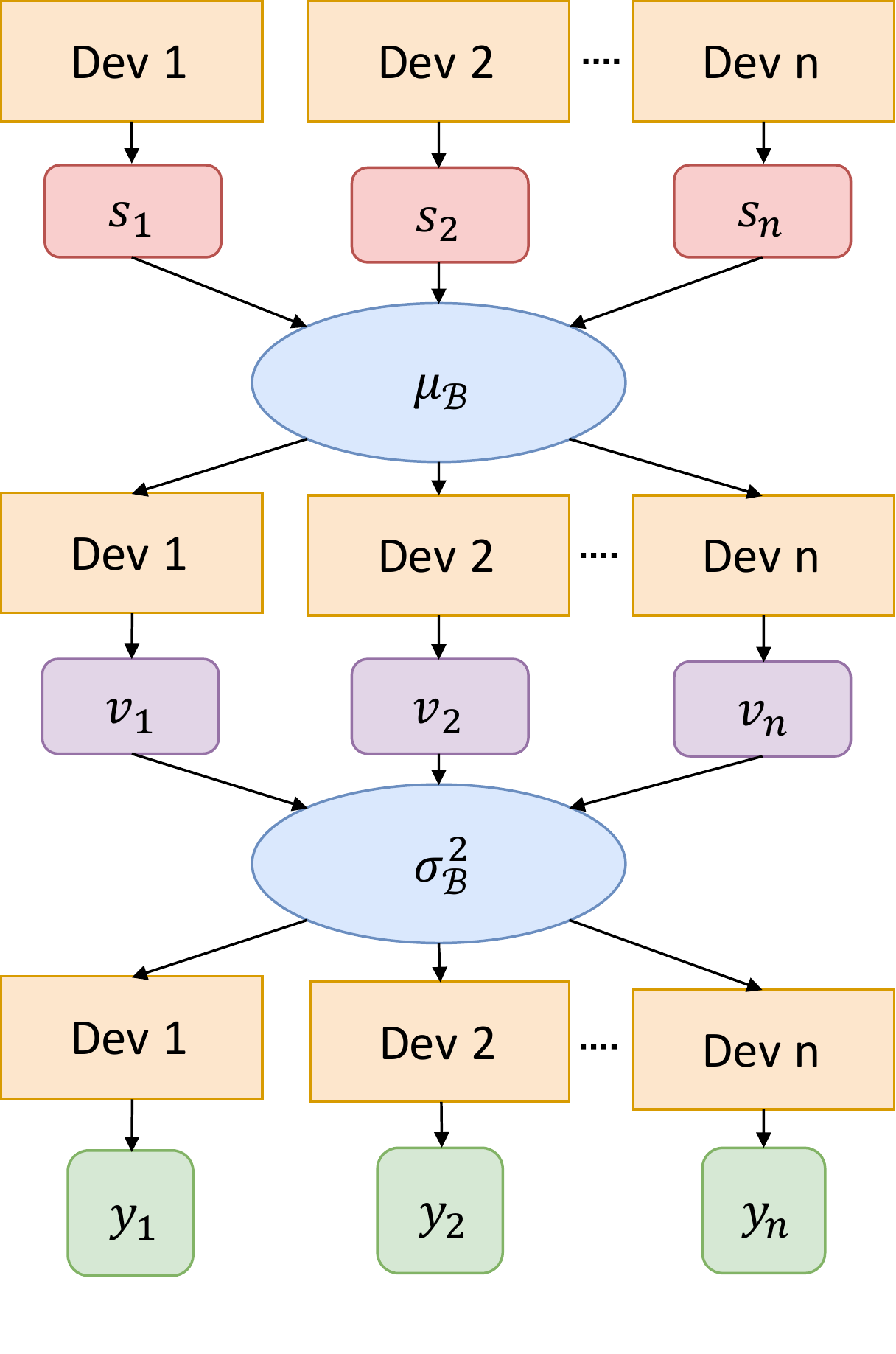}
  \end{center}
  \caption{Implementation of Cross-GPU Batch Normalization. The gray ellipse depicts the synchronization over devices, while the rounded boxes represents paralleled computation of multiple devices.}
  \label{fig:mdbn}
\end{figure}
It is worth noting that the input image of classification network is often $224\times{}224$ or $299\times{}299$, and a single NVIDIA TITAN Xp GPU with 12 Gigabytes memory is enough for 32 or more images. In this way, batch normalization can be computed on each device alone. However, for object detection, a detector needs to handle objects of various scales, thus higher resolution images are needed as its input. In~\cite{Lin_2017_CVPR}, input of size $800\times{}800$ is used, significantly limiting the number of possible samples on one device. Thus, we have to perform batch normalization crossing multiple GPUs to collect sufficient statistics from more samples.

To implement batch normalization across GPUs, we need to compute the aggregated mean/variance statistics over all devices. Most existing deep learning frameworks utilize the BN implementation in cuDNN~\cite{chetlur2014cudnn} that only provides a high-level API without permitting modification of internal statistics. Therefore we need to implement BN in terms of preliminary mathematical expressions and use an ``AllReduce" operation to aggregate the statistics. These fine-grained expressions usually cause significant runtime overhead and the AllReduce operation is missing in most frameworks. 

Our implementation of Cross-GPU Batch Normalization is sketched in Figure~\ref{fig:mdbn}. Given $n$ GPU devices in total, sum value $s_k$ is first computed based on the training examples assigned to the device $k$. By averaging the sum values from all devices, we obtain the mean value $\mu_{\mathcal{B}}$ for current mini-batch. This step requires an AllReduce operation. Then we calculate the variance for each device and get $\sigma^2_{\mathcal{B}}$. After broadcasting $\sigma^2_{\mathcal{B}}$ to each device, we can perform the standard normalization by $y= \gamma \frac{x-\mu_{\mathcal{B}}}{\sqrt{\sigma^2_{\mathcal{B}}+\epsilon}}+ \beta$. Algorithm~\ref{alg:multi_deivce_bn} gives the detailed flow. In our implementation, we use NVIDIA Collective Communication Library (NCCL) to efficiently perform AllReduce operation for receiving and broadcasting.

Note that we only perform BN across GPUs on the same machine. So, we can calculate BN statistics on 16 images if each GPU can hold 2 images. To perform BN on 32 or 64 images, we apply \emph{sub-linear memory}~\cite{chen2016training} to save the GPU memory consumption by slightly compromising the training speed.  

In next section, our experimental results will demonstrate the great impacts of CGBN on both accuracy and convergence. 

\begin{algorithm}
    \caption{Cross-GPU Batch Normalization over a mini-batch $\mathcal{B}$.}
    \label{alg:multi_deivce_bn}
    \begin{algorithmic}[1]
        \REQUIRE
        \begin{tabular}[t]{@{}l}
        Values of input $x$ on multiple devices \\
        in a minibatch: $\mathcal{B} = \bigcup_{i=1}^n \mathcal{B}_i$, $\mathcal{B}_i=\{x_{i_1\ldots i_n}\}$\\
        BN parameters: $\gamma$, $\beta$
        \end{tabular}
        \ENSURE $y = \MDBN{x}$ \\
        \FOR{$i=1,\dots,n$} 
            \STATE {compute the device sum $s_i$ over set $B_i$\\} 
        \ENDFOR
        \STATE{reduce the set $s_{1,\dots,n}$} to minibatch mean $\mu_{\mathcal{B}}$
        \STATE{broadcast $\mu_{\mathcal{B}}$ to each device} 
        \FOR{$i=1,\dots,n$} 
            \STATE {compute the device variance sum $v_i$ over set $B_i$\\} 
        \ENDFOR
        \STATE{reduce the set $v_{1,\dots,n}$} to minibatch variance $\sigma^2_{\mathcal{B}}$
        \STATE{broadcast $\sigma^2_{\mathcal{B}}$ to each device}
        \STATE{compute the output: $y= \gamma \frac{x-\mu_{\mathcal{B}}}{\sqrt{\sigma^2_{\mathcal{B}}+\epsilon}}+ \beta$ over devices}
    \end{algorithmic}
\end{algorithm}

\section{Experiments}

We conduct experiments on COCO Detection Dataset~\cite{lin2014microsoft}, which is split into train, validation, and test, containing 80 categories and over $250,000$ images. We use ResNet-50~\cite{he2016deep} pre-trained on ImageNet~\cite{deng2009imagenet} as the backbone network and Feature Pyramid Network (FPN)~\cite{Lin_2017_CVPR} as the detection framework. We train the detectors over 118,000 training images and evaluate on 5000 validation images. We use the SGD optimizer with momentum 0.9, and adopts the weight decay 0.0001. The base learning rate for mini-batch size 16 is $0.02$. For other settings, the linear scaling rule described in Section~\ref{sec:lr_for_lb} is applied. As for large mini-batch, we use the sublinear memory~\cite{chen2016training} and distributed training to remedy the GPU memory constraints.

We have two training policies in following: 1) \emph{normal}, decreasing the learning rate at epoch 8 and 10 by multiplying scale 0.1, and ending at epoch 11; 2) \emph{long}, decreasing the learning rate at epoch 11 and 14 by multiplying scale 0.1, halving the learning rate at epoch 17, and ending at epoch 18. Unless specified, we use the normal policy.

\subsection{Large mini-batch size, no BN}
\label{sec:without_bn}
We start our study through the different mini-batch size settings, without batch normalization. We conduct the experiments with mini-batch size 16, 32, 64, and 128. For mini-batch sizes 32, we observed that the training has some chances to fail, even we use the warmup strategy. For mini-batch size 64, we are not able to manage the training to converge even with the warmup. We have to lower the learning rate by half to make the training to converge. For mini-batch size 128, the training failed with both warmup and half learning rate. The results on COCO validation set are shown in Table~\ref{tab:bs_no_bn}. We can observe that: 1) mini-batch size 32 achieved a nearly linear acceleration, without loss of accuracy, compared with the baseline using 16; 2) lower learning rate (in mini-batch size 64) results in noticeable accuracy loss; 3) the training is harder or even impossible when the mini-batch size and learning rate are larger, even with the warmup strategy.

    \begin{table}[h]
        \begin{center}
            \begin{tabular}{|l|c|c|}
            \hline
            Mini-Batch size  & mmAP  & Time~(h)\\
            \hline\hline
            16  & 36.2 & 33.2\\
            32  & \textbf{36.4} & \textbf{15.1}\\
            64  & failed & -- \\
            64 (half learning rate) & 36.0 & 7.5 \\
            128 (half learning rate) & failed & -- \\
            \hline
            \end{tabular}
        \end{center}
        \caption{Comparisons of different mini-batch sizes, without BN. }
        \label{tab:bs_no_bn}
    \end{table}

\subsection{Large mini-batch size, with CGBN}
    This part of experiments is trained with batch normalization. Our first key observation is that \emph{all trainings easily converge}, no matter of the mini-batch size, when we combine the warmup strategy and CGBN. This is remarkable because we do not have to worry about the possible loss of accuracy caused by using smaller learning rate. 

	\begin{table}[h]
        \begin{center}
            \begin{tabular}{|c|c|c|c|c|c|}
            \hline
            Batch size & BN size & \# of GPUs & mmAP  & Time(h)\\
            \hline\hline
            16-base & 0 & 8& 36.2 & 33.2 \\
            \hline
            2   & 2 & 2 & 31.5 & 131.2 \\
            4   & 4 & 4 & 34.9 & 91.4 \\
            8   & 8 & 8 & 35.9 & 71.5 \\
            \hline
            16  & 2  & 8 & 31.0 & 45.6 \\
            16  & 16 & 8 & 37.0 & 39.5 \\
            \hline
            32  & 32& 8 & 37.3 & 45.5 \\
            \hline
            64  & 64& 8 & 35.3 & 40.9 \\
            64 & 32& 16 & 37.1 & 19.6\\
            64 & 16& 32 & 37.1 & 11.2\\
            \hline
            128 & 32 & 32& 37.1 & 11.3\\
            128& 16 & 64& 37.0 & 6.5\\
            \hline
            256 & 32 & 64& 37.1 & 7.2\\
            256& 16 & 128& 37.1 & \textbf{4.1}\\
            \hline
            16 (long) & 16 & 8 & 37.7 & 65.2\\
            32 (long) & 32 & 8 & \textbf{37.8} & 60.3\\
            64 (long) & 32 & 16& 37.6 & 30.1\\
            128 (long)& 32 & 32 & 37.6 & 15.8 \\
            256 (long)&32 &  64& 37.7 & 9.4 \\
            256 (long)& 16 & 128& 37.7 & 5.4\\
            \hline
            \end{tabular}
        \end{center}
        \caption{Comparisons of training with different mini-batch sizes, BN sizes (the number of images used for calculating statistics), GPU numbers, and training policies. ``long" means that we apply the long training policy. When the BN size $\geq 32$, the sublinear memory is applied and thus slightly reduces training speed. Overall, the large mini-batch size with BN not only speeds up the training, but also improves the accuracy.}
        \label{tab:bs_with_bn}
    \end{table}	
    The main results are summarized in Table~\ref{tab:bs_with_bn}.  We have the following observations. First, within the growth of mini-batch size, the accuracy almost remains the same level, which is consistently better than the baseline (16-base). In the meanwhile, a larger mini-batch size always leads to a shorter training cycle. For instance, the 256 mini-batch experiment with 128 GPUs finishes the COCO training only in 4.1 hours, which means a $8\times$ acceleration compared to the $33.2$ hours baseline.

    Second, the best BN size (number of images for BN statistics) is 32. With too less images, e.g. 2, 4, or 8, the BN statistics are very inaccurate, thus resulting a worse performance. However, when we increase the size to 64, the accuracy drops. This demonstrates the mismatch between image classification and object detection tasks. 

	\begin{figure}[htbp]
		\centering	
		\includegraphics[scale=0.45]{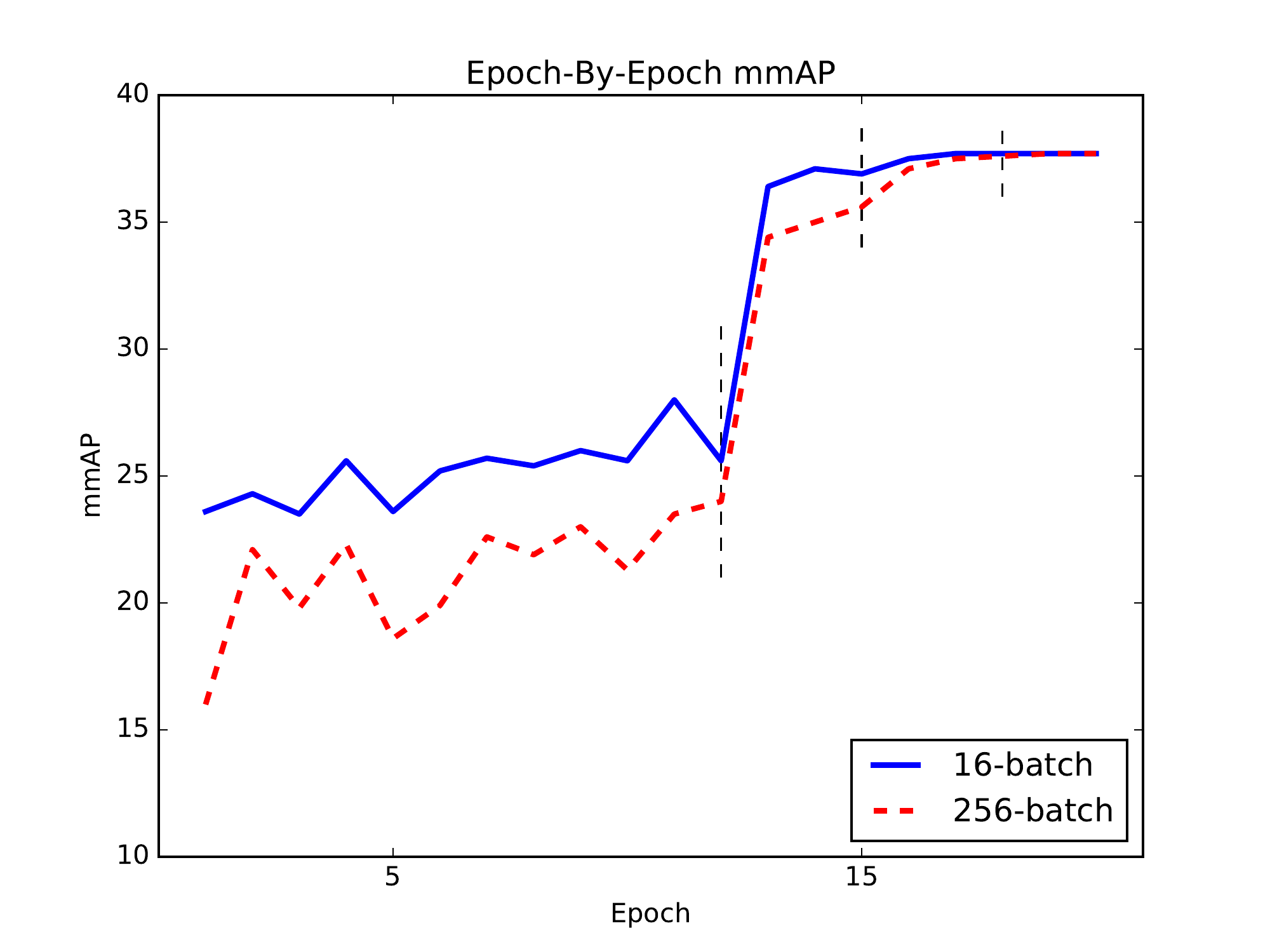}              
		\caption{Validation accuracy of 16 (long) and 256 (long) detectors, using the long training policy. The BN sizes are the same in two detectors. The vertical dashed lines indicate the moments of learning rate decay.}
	\label{fig:epoch_by_epoch_acc}                                                        
	\end{figure}

    Third, in the last part of Table~\ref{tab:bs_with_bn}, we investigate the long training policy. Longer training time slightly boots the accuracy. For example, ``32 (long)" is better that its counterpart (37.8 v.s. 37.3). When the mini-batch size is larger than 16, the final results are very consist, which indicates the true convergence.

    Last, we draw epoch-by-epoch mmAP curves of 16 (long) and 256 (long) in Figure~\ref{fig:epoch_by_epoch_acc}. 256 (long) is worse at early epochs but catches up 16 (long) at the last stage (after second learning rate decay). This observation is different from those in image classification~\cite{goyal2017accurate,you2017100}, where both the accuracy curves and convergent scores are very close between different mini-batch size settings. We leave the understanding of this phenomenon as the future work.

	 \begin{figure*}[t]
      \begin{center}
         \includegraphics[width=0.90\linewidth]{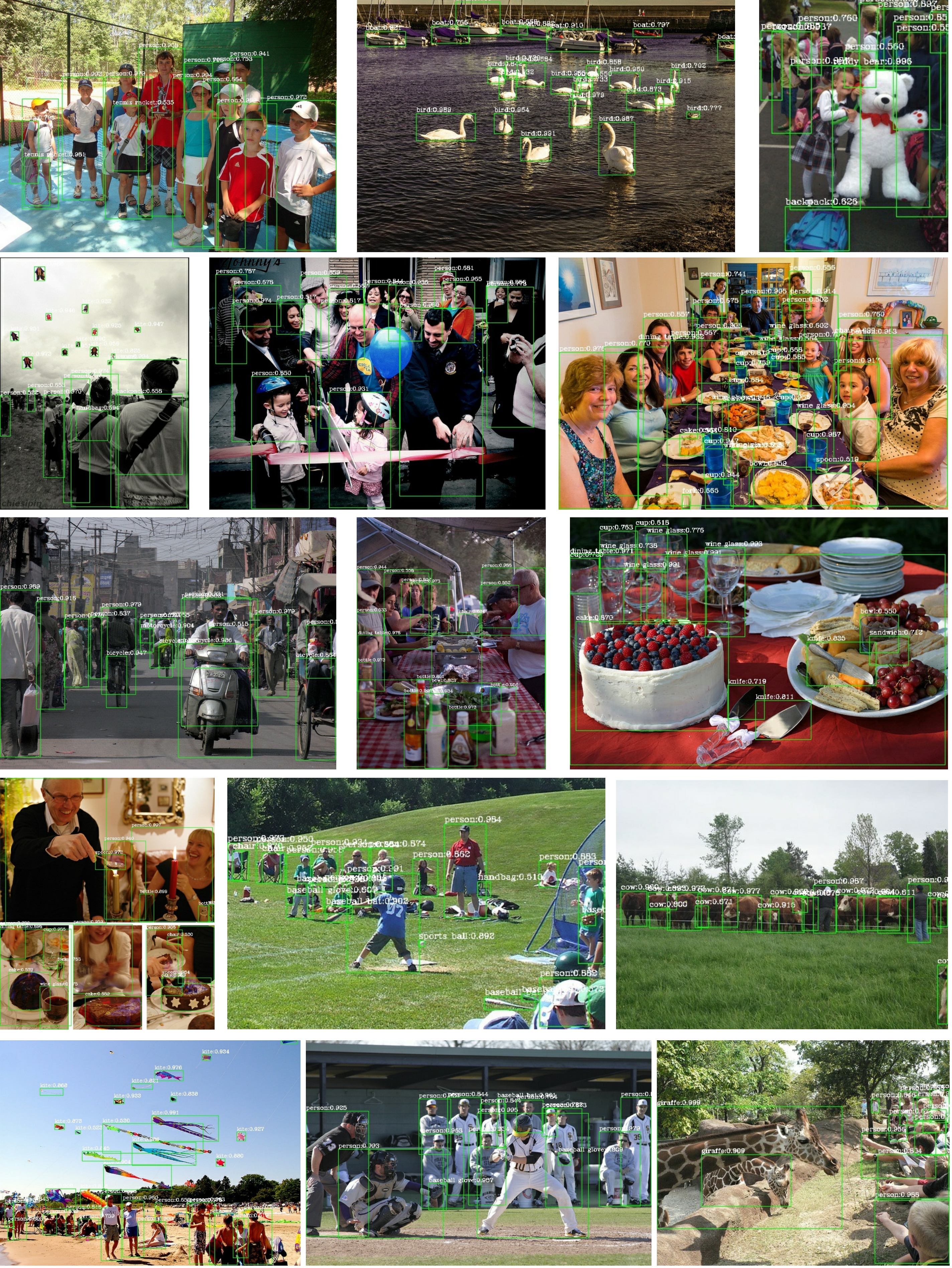}
      \end{center}
      \caption{Illustrative examples for our MegDet on COCO dataset. }
      \label{fig:det_fig_examples}
    \end{figure*}

	\begin{table}[h]
		\begin{center}
			\begin{tabular}{l|c|c}
			\hline
			name & mmAP & \mmAR\\
			\hline
			DANet & 45.7 & 62.7 \\
			Trimps-Soushen+QINIU & 48.0 & 65.4 \\
			bharat\_umd & 48.1 & 64.8\\
			FAIR Mask R-CNN~\cite{He_2017_ICCV} & 50.3 & 66.1\\
			MSRA & 50.4 & 69.0\\
			UCenter & 51.0 & 67.9\\
			\hline
			\textbf{MegDet (Ensemble)} & \textbf{52.5} & \textbf{69.0}\\
			\hline
			\end{tabular}
		\end{center}
		\caption{Result of (enhanced) MegDet on test-dev of COCO dataset.}
		\label{tab:test_dev_on_coco}
	\end{table}

\section{Concluding Remarks}
We have presented a large mini-batch size detector, which achieved better accuracy in much shorter time. This is remarkable because our research cycle has been greatly accelerated. As a result, we have obtained 1st place of COCO 2017 detection challenge. The details are in Appendix.

\section*{Appendix}

Based on our MegDet, we integrate the techniques including OHEM~\cite{shrivastava2016training}, atrous convolution~\cite{yu2015multi,chen2016deeplab}, stronger base models~\cite{Xie_2017_CVPR,hu2017squeeze}, large kernel~\cite{Peng_2017_CVPR}, segmentation supervision~\cite{Mao_2017_CVPR,shrivastava2016contextual}, diverse network structure~\cite{goodfellow2013maxout,ren2017object,szegedy2017inception}, contextual modules~\cite{li2017attentive,gidaris2015object}, ROIAlign~\cite{He_2017_ICCV} and multi-scale training and testing for COCO 2017 Object Detection Challenge. We obtained \textbf{50.5} mmAP on validation set, and \textbf{50.6} mmAP on the test-dev. The ensemble of four detectors finally achieved \textbf{52.5}.  Table~\ref{tab:test_dev_on_coco} summarizes the entries from the leaderboard of COCO 2017 Challenge. Figure~\ref{fig:det_fig_examples} gives some exemplar results.

	{\small
\bibliographystyle{ieee}
\bibliography{egbib}
}

\end{document}